\documentclass[10pt,twocolumn,letterpaper]{article}

\usepackage[pagenumbers]{cvpr} %

\usepackage[accsupp]{axessibility}  %

\usepackage{graphicx}
\usepackage{amsmath}
\usepackage{amssymb}
\usepackage{booktabs}
\usepackage{enumerate}
\usepackage{multirow}
\usepackage{float}

\usepackage[numbers,sort&compress]{natbib}
\usepackage{cite} 
\usepackage{times}
\usepackage{epsfig}
\usepackage{graphicx}
\usepackage{amsmath}
\usepackage{amssymb}
\usepackage{enumerate}
\usepackage{booktabs}
\usepackage{multirow}
\usepackage[table,xcdraw]{xcolor}
\usepackage{booktabs}
\usepackage{multirow}
\usepackage[normalem]{ulem}
\useunder{\uline}{\ul}{}

\usepackage{times}
\usepackage{epsfig}
\usepackage{graphicx}
\usepackage{amsmath}
\usepackage{amssymb}
\usepackage{tabulary}
\usepackage{multirow}
\usepackage{silence}
\usepackage{caption}
\usepackage{xfrac}
\usepackage{float}
\usepackage[pagebackref=true,breaklinks=true,letterpaper=true,colorlinks,bookmarks=false]{hyperref}

\usepackage{algpseudocode}
\usepackage[]{algorithm2e}

\makeatletter
\renewcommand\paragraph{\@startsection{paragraph}{4}{\z@}%
                                    {0.5ex \@plus1ex \@minus.1ex}%
                                    {-1em}%
                                    {\normalfont\normalsize\bfseries}}

\newcommand{\vspaceaftersection}{-0.18cm}
\newcommand{\vspacebeforesection}{-0.18cm}

\newcommand{\Ours}{{\textit{GART}}}

\makeatother

\usepackage[capitalize]{cleveref}
\crefname{section}{Sec.}{Secs.}
\Crefname{section}{Section}{Sections}
\Crefname{table}{Table}{Tables}
\crefname{table}{Tab.}{Tabs.}

\usepackage{enumitem}
\usepackage[numbers]{natbib}
\usepackage{colortbl}

\usepackage{multicol}

\usepackage{color}
\usepackage{xcolor}
\definecolor{turquoise}{cmyk}{0.65,0,0.1,0.3}
\definecolor{purple}{rgb}{0.65,0,0.65}
\definecolor{dark_green}{rgb}{0, 0.5, 0}
\definecolor{orange}{rgb}{0.8, 0.6, 0.2}
\definecolor{red}{rgb}{0.8, 0.2, 0.2}
\definecolor{darkred}{rgb}{0.6, 0.1, 0.05}
\definecolor{blueish}{rgb}{0.0, 0.3, .6}
\definecolor{light_gray}{rgb}{0.7, 0.7, .7}
\definecolor{pink}{rgb}{1, 0, 1}
\definecolor{greyblue}{rgb}{0.25, 0.25, 1}
\definecolor{tab_blue}{HTML}{1f77b4}
\definecolor{tab_orange}{HTML}{ff7f0e}
\definecolor{LightRed}{rgb}{0.99,0.89,0.89}

\definecolor{mesh_misty_rose}{HTML}{e6aaa3}
\definecolor{mesh_yellow}{HTML}{ffba00}

\def\cvprPaperID{XXXXX}
\def\confName{CVPR}
\def\confYear{2024}
\begin{document}

\title{\emph{GART}: \underline{Ga}ussian \underline{Art}iculated \underline{T}emplate Models}
\author{
        Jiahui Lei\textsuperscript{1} \quad Yufu Wang\textsuperscript{1} \quad Georgios Pavlakos\textsuperscript{2} \quad  Lingjie Liu\textsuperscript{1} \quad Kostas Daniilidis\textsuperscript{1,3}\\
        $^1$ University of Pennsylvania \qquad
        $^2$ UC Berkeley \qquad
        $^3$ Archimedes, Athena RC\\
        {\tt\small \{leijh, yufu, lingjie.liu, kostas\}@cis.upenn.edu, pavlakos@berkeley.edu} 
    }

\twocolumn[\maketitle\vspace{-0.5em}\begin{center}
\vspace{-3em}
\includegraphics[width=1\linewidth]{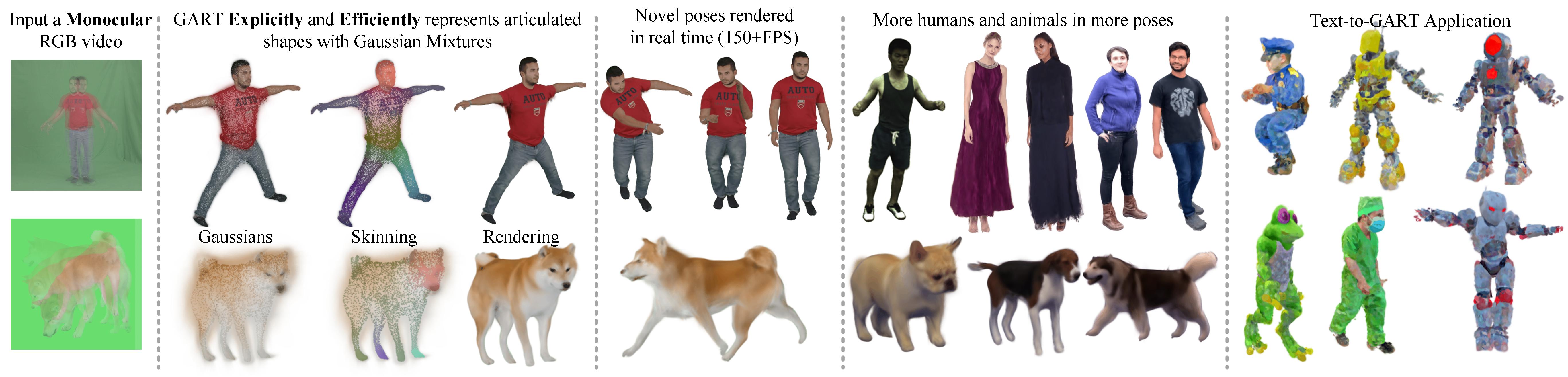}
\end{center}\vspace{-2em}

\captionof{figure}{
We propose {\Ours}, an explicit, efficient, and expressive model for articulated object capturing and rendering from monocular videos.
Code available on the project page
\url{https://www.cis.upenn.edu/~leijh/projects/gart/}.
}
\label{fig:teaser}
\bigbreak]

\begin{abstract}
We introduce Gaussian Articulated Template Model (\Ours), an \textbf{explicit}, \textbf{efficient}, and \textbf{expressive} representation for non-rigid articulated 
subject capturing and rendering from monocular videos. 
{\Ours} utilizes a mixture of moving 3D Gaussians to explicitly approximate a deformable subject's geometry and appearance. 
It takes advantage of a categorical template model prior (SMPL, SMAL, etc.) with learnable forward skinning while further generalizing to more complex non-rigid deformations with novel latent bones.
{\Ours} can be reconstructed via differentiable rendering from monocular videos in seconds or minutes and rendered in novel poses faster than 150fps.  
\end{abstract}

\vspace{\vspacebeforesection}
\section{Introduction}
\vspace{\vspaceaftersection}
\label{sec:intro}
Humans and animals are the most common deformable entities in real-world dynamic scenes, hence the plethora of approaches to modeling their geometry, appearance, and motion. 
This paper studies how to represent and reconstruct such deformable subjects from \textbf{monocular} videos. Since they share category-level structures, morphable template models are developed and widely applied, such as SMPL~\cite{smpl} for humans and SMAL~\cite{smal} for quadrupedal animals. While they are useful for pose estimation, categorical template models cannot capture detailed appearance and geometry during a variety of deformations.

Recent studies proposed to address this problem by building additional implicit representations on templates in order to model geometry deformation and appearance.
Most of these representations are based on neural fields~\cite{mildenhall2021nerf, park2019deepsdf, mescheder2019occupancy,xie2022neural}. {\bf Implicit} representations enhance quality but suffer from slow rendering because of costly query operations. Animating neural fields is challenging and requires specialized forward or backward skinning~\cite{snarf,fastsnarf}. Moreover, these methods usually depend on accurate template pose estimation since they can easily create artifacts in the empty space when the stereo is wrong. On the contrary,  {\bf explicit} representations~\cite{habermann2021real,peopelsnap,zheng2023pointavatar} are efficient to render, easy to deform, and more robust to pose estimation errors because of the deformation-based optimization process. However,  explicit representations often have sub-optimal quality and are restricted by fixed mesh topology~\cite{peopelsnap}, constrained by using too many points~\cite{zheng2023pointavatar}, or heavily rely on the multi-view studio camera system~\cite{habermann2021real}.

Our main insight into the articulation modeling was that \textbf{an explicit approximation for the implicit radiance field} would overcome the weaknesses of both worlds. We propose Gaussian Articulated Template Models ({\Ours), a new renderable representation for non-rigid articulated subjects.
{\Ours} takes advantage of classical template models by using its kinematic skeleton and models the detailed appearance via a Gaussian Mixture Model (GMM) in the canonical space that approximates the underlying radiance field (Sec.~\ref{sec:shape_gmm}). Because a GMM does not have a fixed topology and each component can smoothly approximate a neighborhood, {\Ours} is as expressive as NeRFs while maintaining simplicity and interpretability.

As an explicit representation, {\Ours} can be animated via forward skinning similar to template meshes. However, the predefined skeleton cannot capture the movements of loose clothes such as long dresses. We address this challenge with a novel latent bone approach, where a set of unobserved latent bones, as well as their skinning weights that drive the additional deformation, can be simultaneously learned from a monocular video (Sec.~\ref{sec:motion}).
Another challenge of the GMM approximation is the lack of local smoothness compared to neural fields, which impacts the reconstruction quality when the input views are sparse or the input human poses are noisy. We introduce smoothness priors for modeling articulated subjects to adapt {\Ours} for monocular reconstruction (Sec.~\ref{sec:reconstruct}). 

To capture an articulated subject from monocular video, we initialize {\Ours} with the estimated template, and render the GMM with 3D Gaussian Splatting~\cite{zwicker2002ewa, kerbl20233d} to reconstruct each frame. The optimization process gradually updates each Gaussian parameter and operates like a deformation-based approach that behaves more robustly under errors in the template pose estimations. 
With the explicit, efficient, and expressive {\Ours}, we are able to reconstruct a human avatar from a monocular video in $30$ seconds and render it with resolution $540\times 540$ at $150+$ FPS on a laptop, to our current knowledge, faster than any state-of-the-art NeRF-based human rendering methods. Furthermore, we use {\Ours} as a general framework to reconstruct animals from monocular videos in the wild with higher fidelity than previous mesh-based approaches. 

In summary, our main contributions are: 1) {\Ours}, a general and explicit representation for non-rigid articulated subjects, which approximates the radiance field of the canonical shape and appearance with a Gaussian Mixture Model; 2) {\Ours} can be efficiently animated via learnable forward skinning and can capture challenging deformations such as loose clothes on humans via a novel latent bones approach;
3) Our experiments demonstrate that GART achieves SoTA performance in monocular human reconstruction and rendering on various datasets with the best training and inference efficiency and produces high-quality animal reconstruction from monocular videos in the wild.
\section{Related Work}
\vspace{\vspaceaftersection}
\label{sec:related}
\textbf{3D Human Reconstruction}. Reconstructing 3D humans from monocular observations is a difficult task due to depth ambiguity. Parametric template models~\cite{smpl,smplx, flame, mano} provide a strong prior of the human body and are key in the recent advances of monocular 3D human pose and shape reconstruction~\cite{kanazawa2018end, kolotouros2019learning, kocabas2020vibe, sun2021monocular, sun2023trace, goel2023humans, wang2023refit}. The explicit and predefined topology of parametric meshes, however, cannot capture personalized appearance such as texture, hair, and clothing~\cite{peopelsnap, alldieck2018video, habermann2019livecap, habermann2020deepcap, habermann2021real}. 
To address this issue, recent studies~\cite{jiang2023instantavatar, guo2023vid2avatar,he2021arch++,saito2020pifuhd, xiu2022icon, instant_nvr, jiang2022selfrecon, zielonka2023instant, gafni2021dynamic,yu2023monohuman,liu2021neural,peng2021animatable,chen2021animatable,xu2021hnerf, su2021anerf, 2021narf, zhang2021stnerf, kwon2021nhp, NNA, zheng2022structured, ARAH:2022:ECCV, weng2022humannerf,li2022tava,su2022danbo,peng2023implicit,jiang2022neuman,kwon2023deliffas,HDHumans,snarf,fastsnarf} propose to use neural representations, such as NeRF, to capture high-fidelity humans from multiple views or videos. To reconstruct dynamic humans, neural representations are combined with parametric models to disentangle pose and shape~\cite{peng2021neural, liu2021neural, weng2022humannerf}. Appearance can be modeled in the canonical space and then posed by the articulated template~\cite{snarf, fastsnarf}. These hybrid approaches allow re-animation of the captured avatar and demonstrate high flexibility to model personalized details. However, one drawback is their inefficiency in querying and rendering. 
Our proposed {\Ours} similarly utilizes parametric templates to model the human body articulation. But unlike the above neural representations, the appearance is represented by 3D Gaussians~\cite{kerbl20233d, Zielonka2023Drivable3D, zwicker2002ewa} that are efficient to render. Additionally, the explicitness of 3D Gaussians allows us to design simple deformation and regularization rules.

\textbf{3D Animal Reconstruction}. Similar to the modeling of humans, parametric models have been proposed for different animals~\cite{smal, biggs2020left, ruegg2023bite, li2021hsmal, badger20203d} and can be fitted to images and videos~\cite{zuffi2019three, biggs2019creatures}. Novel instances or more species can be captured  with limited fidelity by deforming the template with image guidance~\cite{zuffi2018lions, wang2021birds}. As template models are expensive to create for diverse animals, model-free approaches learn animal shapes by deforming a sphere~\cite{cashman2012shape, goel2020shape, wu2023dove, tulsiani2020implicit}. Recent approaches aim to build articulated models directly from videos as animatable neural fields~\cite{wu2023magicpony, yang2022banmo, yang2023reconstructing}. High-quality neural capture of animals has been demonstrated with a multi-view RGB and Vicon camera array system~\cite{luo2022artemis}.
However, unlike all these methods, {\Ours} robustly builds detailed 3D Gaussians upon D-SMAL~\cite{ruegg2023bite} templates and can capture diverse dog species from in-the-wild monocular videos.

\textbf{3D Gaussian Splatting}.
The key technique of the above-mentioned reconstruction now lies in differentiable rendering, where meshes~\cite{laine2020modular, liu2019soft}, points/surfels~\cite{yifan2019differentiable,lassner2021pulsar,zhang2022differentiable,xu20234k4d} and NeRFs~\cite{mildenhall2021nerf,muller2022instant,li2022nerfacc} have been widely used. Recent progress in differentiable rendering revives the classical EWA volume splatting~\cite{zwicker2002ewa}, in which 3D Gaussians are used to approximate the underlying radiance field and achieve high efficiency and fidelity~\cite{kerbl20233d} via splatting-based rendering.
3D-GS~\cite{kerbl20233d,keselman2023flexible} techniques have been recently applied to modeling general dynamic scenes~\cite{yang2023deformable,luiten2023dynamic,wu20234d,yang2023real} where the scenes do not have specific structures (i.e. articulation), and have been applied to 3D generation~\cite{yi2023gaussiandreamer,tang2023dreamgaussian,chen2023text}.
\vspace{\vspacebeforesection}
\section{Method}
\vspace{\vspaceaftersection}

\begin{figure*}[h!]
\centering
   \includegraphics[width=1.0\textwidth]{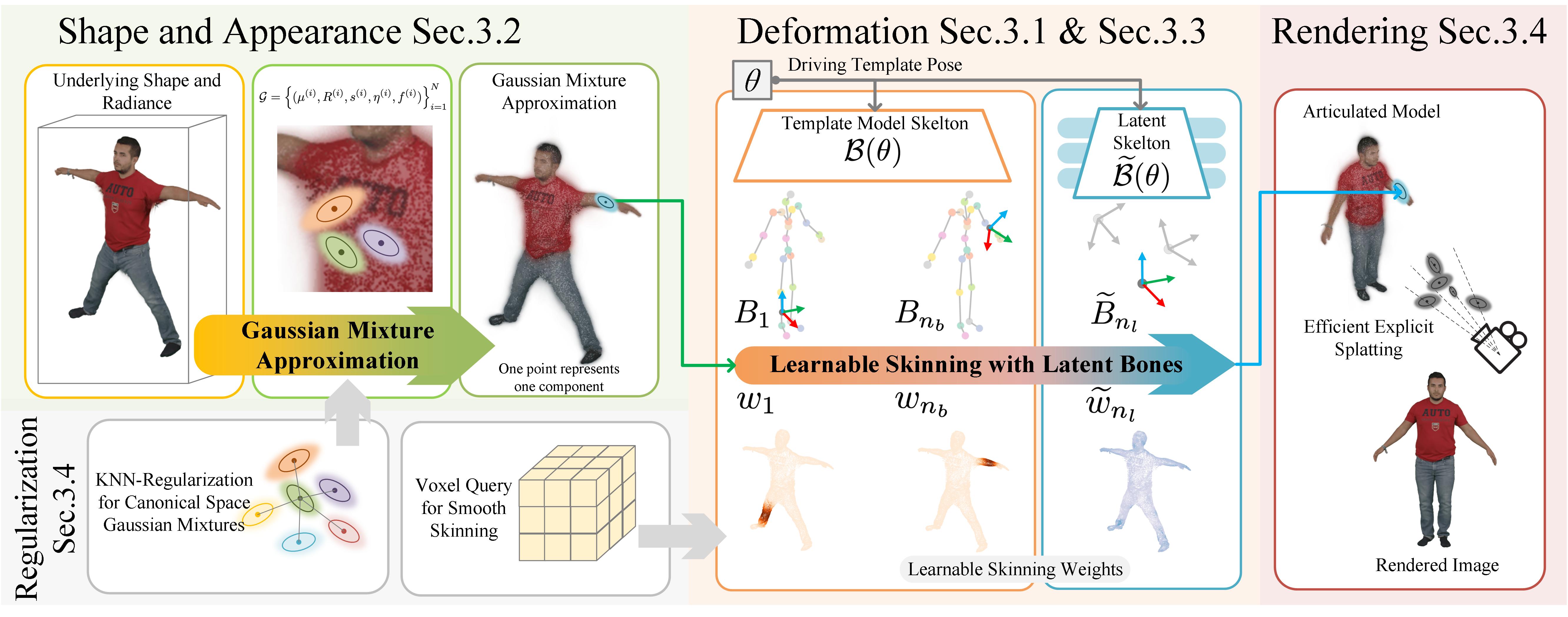}
    \caption{\textbf{Overview}:
    Left-top: {\Ours} represents the shape and appearance of articulated subjects in canonical space with Gaussian Mixtures (Sec.~\ref{sec:shape_gmm}). Middle: Such explicit approximation can be efficiently deformed with learnable skinning and a novel latent bone approach, capturing challenging deformations (Sec.~\ref{sec:motion}). Right: The articulated model can be efficiently rendered via Gaussian Splatting~\cite{zwicker2002ewa,kerbl20233d} (Sec.~\ref{sec:reconstruct}). Left-bottom: Several smoothness regularizations are injected into {\Ours} to constrain the point-based representation (Sec.~\ref{sec:reconstruct}).
    }
    \label{fig:main}
\end{figure*}

\subsection{Template Prior}
\label{sec:template}
\vspace{\vspaceaftersection}
Different from capturing deforming subjects from multi-view systems~\cite{habermann2019livecap, habermann2020deepcap} in a studio, the capture from monocular videos is extremely challenging, and many studies leverage category-level template models as a strong prior to associating and accumulating information across time for monocular reconstruction of humans and animals. 
These templates include the SMPL~\cite{smpl,smplx} family for humans and SMAL~\cite{smal, ruegg2023bite} for animals.
Typically, a template model consists of three components:
\begin{equation}
    (\mathcal M, \mathcal B, \mathcal W).
    \label{eq:tempalte_triplet}
\end{equation}
The template mesh $\mathcal M=(\mathcal V_c, \mathcal F)$ is defined in the canonical space to model the shape of the subjects. Predefined motion structure (skeleton) $\mathcal B$ of the category with $n_b$ joints can return a set of rigid bone transformations based on the driven pose $\theta$:
\begin{equation}
    [B_1, B_2, \ldots, B_{n_b}] = \mathcal B(\theta),
    \label{eq:bones}
\end{equation}
where $B_i \in SE(3)$ represents the rigid transformation that moves the canonical joint coordinate frame to the articulated one. The surface point $x_c$ in the canonical space can be deformed to the articulated space via the linear blend skinning (LBS):
\begin{equation}
    x = \left(\sum_{k=1}^{n_b} \mathcal W_k(x_c) B_k \right)x_c,
    \label{eq:forward_skinning}
\end{equation}
where $\mathcal W_k(x_c) \in \mathbb R$ is querying the predefined skinning weight in canonical space. Usually, $\mathcal W$ can be predefined in the full $\mathbb R^3$ space by diffusing the mesh skinning weights.

\vspace{\vspacebeforesection}
\subsection{Shape Appearance Representation with GMM} 
\vspace{\vspaceaftersection}
\label{sec:shape_gmm}

Gaussian Articulated Template (\Ours) is a representation for deformable articulated subjects that combines the advantages of implicit and explicit representations. Inspired by recent progress in static scene rendering~\cite{kerbl20233d} and classical point-based graphics~\cite{zwicker2002ewa}, we propose to \emph{use 3D Gaussian Mixture Models (GMM) to \underline{explicitly} approximate the \underline{implicit} underlying radiance field in the canonical space}. The $i$th component in the GMM is parameterized by a 3D mean $\mu^{(i)} \in \mathbb R^3$, a 3D rotation $R^{(i)} \in SO(3)$, anisotropic 3-dimensional scaling factors $s^{(i)} \in \mathbb R^{3+}$, an opacity factor $\eta^{(i)} \in (0,1]$, and the color radiance function encoded by Spherical harmonics $f^{(i)} \in \mathbb R^{C}$. Given a query position $x_c$ in canonical space, the density contribution of $i$th component is:
\begin{equation}
    \sigma^{(i)}(x_c) =\eta \exp{\left( -\frac{1}{2 }(x_c-\mu)^T \varSigma^{-1} (x_c-\mu)\right)},
    \label{eq:3dgs_density}
\end{equation}
where the covariance is $\varSigma=R \text{diag}(s^2)R^T$, and we omit the index $i$. The color contribution of this component is:
\begin{equation}
    c^{(i)}(x_c, d) = \sigma(x_c) \text{sph}(R^Td,f),
    \label{eq:3dgs_color}
\end{equation}
where $\text{sph}(R^Td,f)$ means evaluating the spherical harmonics with coefficient $f$ at the local frame direction $R^Td$ given the global query viewing direction $d$. The total radiance field is the summation of all $N$ components:
\begin{equation}
    \sigma(x_c) = \sum_{i=1}^N\sigma^{(i)}(x_c), \quad c(x_c, d) = \sum_{i=1}^N c^{(i)}(x_c, d).
    \label{eq:3dgs_mixture}
\end{equation}
Eq.~\ref{eq:3dgs_mixture} explicitly represents the canonical geometry and appearance by a list of Gaussian component parameters, which can be written as:
\begin{equation}
    \mathcal G =\left\{ (\mu^{(i)}, R^{(i)}, s^{(i)}, \eta^{(i)}, f^{(i)}) \right\}_{i=1}^N.
\end{equation}
In \Ours, $\mathcal G$ replaces the $\mathcal M$ in Eq.~\ref{eq:tempalte_triplet} triplet. Note that during optimization (Sec.~\ref{sec:reconstruct}), each component can move independently and is not constrained by a fixed topology like a mesh, which makes {\Ours} highly flexible.

\vspace{\vspacebeforesection}
\subsection{Motion Representation with Forward Skinning}
\label{sec:motion}
\vspace{\vspaceaftersection}

\paragraph{Learnable Forward Skinning}
One key advantage of using $\mathcal G$ is the simple and explicit deformation modeling. Since the predefined category-level skinning prior $\mathcal W$ from the template may not reflect the actual instance deformation, we assign each Gaussian component a learnable skinning correction:
\begin{equation}
    \widehat{\mathcal W}(\mu^{(i)}) = \mathcal W (\mu^{(i)}) + \Delta w^{(i)},
    \label{eq:learnable_skinning}
\end{equation}
where $\Delta w^{(i)}\in \mathbb R^{n_b}$ is the learnable skinning of the $i$th Gaussian. In \Ours, learnable $\widehat{\mathcal W}$ replaces the $\mathcal W$ in Eq.~\ref{eq:tempalte_triplet}.
Given a pose $\theta$, the articulation transformation $A^{(i)}$ for the $i$th Gaussian is:
\begin{equation}
    A^{(i)} = \sum_{k=1}^{n_b} \widehat{\mathcal W_k}(\mu^{(i)}) B_k,
    \label{eq:gauss_transformation}
\end{equation}
and the Gaussian center and rotation are articulated to:
\begin{equation}
    \mu_{\text{art}}^{(i)} = A^{(i)}_{\text{rot}}\mu^{(i)} + A_t^{(i)}, \quad R_{\text{art}}^{(i)} = A_{\text{rot}}R^{(i)},
    \label{eq:articualted_mu_R}
\end{equation}
where $A_{\text{rot}}^{(i)}$, $A_t^{(i)}$ are the left top $3\times 3$ and the right $3 \times 1$ block of $A^{(i)}$. Note that $A^{(i)}$ may not be an $SE(3)$ transformation anymore. But the transformed $R_a^{(i)}$ can still be used to compose a covariance as in Eq.~\ref{eq:3dgs_density}, and the articulated radiance field can be directly obtained from Eq.~\ref{eq:3dgs_mixture} by using the articulated mean and covariance as in Eq.~\ref{eq:articualted_mu_R}. This forward skinning enables {\Ours} to model motion efficiently and to avoid backward skinning root-finding~\cite{snarf, fastsnarf}, which is used in other implicit representations.

\paragraph{Latent Bones and Flexible Deformation}
Person-agnostic human models such as SMPL have a predefined skeleton $\mathcal B$ that models the human body motion well but cannot capture the movement of loose clothing. 
Our goal is to find a simple approximation for the clothing motion that can be captured from a monocular video. Our insight is that the deformation of an articulated subject can be seen as driven by the $n_b$ predefined bones plus $n_l$ unknown latent bones. We can represent the latent bone transformations as a function of the pose $\theta$:
\begin{equation}
    [\widetilde B_{1},\ldots, \widetilde B_{n_l}] = \widetilde{\mathcal{B}}(\theta)
    \label{eq:add_bones}
\end{equation}
where $\widetilde B_{i} \in SE(3)$ and $\tilde{\mathcal{B}}(\theta)$ can be parameterized with an MLP or a per-frame optimizable table. Similarly, we can learn the latent bone skinning weight for each Gaussian $\widetilde{\mathcal{W}}(\mu)\in \mathbb R^{n_l}$ during training. With the addition of latent bones, the forward skinning from Eq.~\ref{eq:gauss_transformation} became
\begin{equation}
    A^{(i)} = \sum_{k=1}^{n_b} \widehat{\mathcal W_k}(\mu^{(i)}) B_k + \sum_{q=1}^{n_l} \widetilde{\mathcal W_q}(\mu^{(i)}) \widetilde B_q.
    \label{eq:gauss_transformation_flex}
\end{equation}
Note this deformation model is computationally efficient and compact since the transformations $\mathcal B$, $\widetilde{\mathcal{B}}$ are globally shared across all Gaussians. 
\paragraph{Summary} Now, we fully introduced \Ours:
\begin{equation}
    (\mathcal G, \mathcal B, \widehat{\mathcal W}, \widetilde{\mathcal B}, \widetilde{\mathcal W}),
\end{equation}
which explicitly approximates the canonical shape and appearance with learnable GMM $\mathcal G$, and compactly represents the forward deformation with prior skeleton and learnable latent bones $\mathcal B,\widetilde{\mathcal B}$, and their learnable skinning weights $\widehat{\mathcal W}, \widetilde{\mathcal W}$.
Given a pose $\theta$, using Eq.~\ref{eq:articualted_mu_R},\ref{eq:gauss_transformation_flex}, the articulated radiance field approximation is:
\begin{align}
    \begin{split}
        \mathcal G_{\text{art}}(\theta) &= \left\{ (\mu_{\text{art}}^{(i)}, R_\text{art}^{(i)}, s^{(i)}, \eta^{(i)}, f^{(i)}) \right\}_{i=1}^N\\
        \mu_{\text{art}}^{(i)} &= A^{(i)}_{\text{rot}}\mu^{(i)} + A_t^{(i)}, \quad R_{\text{art}}^{(i)} = A_{\text{rot}}R^{(i)}\\
        A^{(i)} &= \sum_{k=1}^{n_b} \widehat{\mathcal W_k}(\mu^{(i)}) B_k(\theta) + \sum_{q=1}^{n_l} \widetilde{\mathcal W_q}(\mu^{(i)}) \widetilde B_q(\theta)
    \end{split}
\end{align}

\vspace{\vspacebeforesection}
\subsection{Reconstruct {\Ours} from Monocular Videos}
\vspace{\vspaceaftersection}
\label{sec:reconstruct}

\paragraph{Differentiable Rendering with Splatting}
Given a perspective projection $\pi(x; E, K)$ where $E$ is the camera extrinsics and $K$ the intrinsics matrix, the projection of a 3D Gaussian can be approximately treated as a 2D Gaussian with mean and covariance:
\begin{equation}
    \mu_{2D} = \pi(\mu; E,K) ; \ \varSigma_{2D} = J E \varSigma E^T J^T,
    \label{eq:3dgs_approx}
\end{equation}
where $J$ is the Jacobian of the perspective projection, see equations~(26-31) in \cite{zwicker2002ewa}. With the preservation of the Gaussians through projection, we can efficiently splat and approximate the volume rendering with sorted color accumulation~\cite{zwicker2002ewa}:
\begin{align}\begin{split}
  I(u,d) &= \sum_{i=1}^{N} \alpha^{(i)} \text{sph}(R^Td,f^{(i)}) \prod_{j=1}^{i-1} (1-\alpha^{(j)})\\
  \alpha^{(i)} &= G_{2D}(u|\eta^{(i)}, \mu_{2D}^{(i)},\varSigma_{2D}^{(i)}),
  \label{eq:3dgs_rendering}
\end{split}\end{align}
where the index is sorted along the depth direction, the querying pixel coordinate is $u$, the viewing direction in the world frame is $d$, and $G_{2D}$ is evaluating the 2D Gaussian density similar to Eq.~\ref{eq:3dgs_density}. 
Eq.~\ref{eq:3dgs_rendering} is differentiable~\cite{kerbl20233d} and can provide supervision from 2D observations to update all Gaussian parameters, and we refer the readers to \cite{zwicker2002ewa,kerbl20233d} for more details.

\paragraph{Optimization}
Given a set of $M$ images from a monocular video and the estimated poses of the template $\{(I^*_1, \theta_1), \ldots, (I^*_M, \theta_M)\}$, we optimize $\mathcal G, \widehat{\mathcal W}, \widetilde{\mathcal B}, \widetilde{\mathcal W}$ as well as refine $\theta$ by comparing the rendered image of  $\mathcal G_{\text{art}}(\theta)$ with the ground truth images. We initialize $\mathcal G$ on the template mesh and follow the densify-and-prune strategy from 3D-GS~\cite{kerbl20233d} during optimization.
Denote the rendered image of {\Ours} as $\hat I(\mathcal G_{\text{art}}(\theta))$, the training loss is:
\begin{equation}
    L = L_1(\hat I, I^*) + \lambda_{\text{SSIM}} L_{\text{SSIM}}(\hat I, I^*) + L_{\text{reg}},
    \label{eq:total_loss}
\end{equation}
where $\lambda$ is the loss weight, and $L_{\text{reg}}$ is introduced below.
\paragraph{Regularization} The flexible nature of the 3D Gaussians in {\Ours} can be under-constrained when the 2D observation is sparse. 3D Gaussian mixture does not have the smoothness induced by MLPs as in NeRFs, which often leads to artifacts in unobserved spaces. Inspired by \cite{snarf, fastsnarf}, the learnable skinning weights $\widehat{\mathcal W},\widetilde{\mathcal W} $ should be spatially smooth, so we distill them to a coarse voxel grid, and the per-Gaussian skinning $\Delta w^{(i)}, \widetilde{\mathcal W}(\mu^{(i)})$ in Eq.~\ref{eq:learnable_skinning},\ref{eq:gauss_transformation_flex} are tri-linearly interpolated at $\mu^{(i)}$ from the voxel grid.
We further regularize the variation of the Gaussian attributes in the KNN neighborhood of $\mu$, which leads to:
\begin{equation}
    L^{(i)}_{STD} = \sum_{attr \in \{R,s,\eta, f,\widehat{\mathcal W}, \widetilde{\mathcal W}\}} \lambda_{attr} \text{STD}_{i\in KNN(\mu^{(i)})}(attr^{(i)}),
    \label{eq:knn_reg}
\end{equation}
where $STD$ is the standard deviation. 
Additionally, we encourage the fitting to make small changes from the original motion structure to further exploit the template model prior knowledge and to encourage small Gaussians since the non-rigid subject is approximated by piece-wise rigid moving Gaussians, which leads to:
\begin{equation}
    L^{(i)}_{norm} = \lambda_{ \widehat{\mathcal W}} \| \Delta w^{(i)} \|_2 + \lambda_{ \widetilde{\mathcal W}} \|  \widetilde{\mathcal W}(\mu^{(i)}) \|_2 + \lambda_s \|s^{(i)}\|_{\infty}.
\end{equation}
The total regularization loss is:
\begin{equation}
    L_{reg} = \frac{1}{N}\sum_{i=1}^N L^{(i)}_{STD} + L^{(i)}_{norm}.
\end{equation}

\paragraph{Inference}
During inference, all the attributes of {\Ours} are explicitly stored per Gaussian (no voxel grid query is needed). With our efficient modeling of appearance and motion, rendering an articulated subject is as fast as rendering a static scene. The inference FPS is more than $150$ on People-Snapshot~\cite{peopelsnap} at resolution $540\times 540$.

\vspace{\vspacebeforesection}
\section{Experiments}
\vspace{\vspaceaftersection}

\subsection{Comparison on Human Rendering}
\vspace{\vspaceaftersection}
In this section, we verify \Ours's effectiveness, efficiency,
and expressiveness in monocular human reconstruction and view synthesis.
We use SMPL~\cite{smpl} as the template, and the input during training is a monocular RGB video with an estimated SMPL pose at each video frame. The evaluation during testing is the novel view synthesis with the PSNR, SSIM, and LPIPS metrics.

The SoTA baselines in this task are the recent efficient NeRF-based human rendering methods Instant-Avatar~\cite{jiang2023instantavatar} and Instant-NVR~\cite{instant_nvr}, which demonstrate better fidelity than classical mesh-based representations~\cite{peopelsnap}. 
Instant-Avatar uses instant-NGP~\cite{muller2022instant} in the canonical space and utilizes Fast-SNARF~\cite{fastsnarf}, a highly tailored GPU solver for fast backward skinning root finding, to model the deformation. It also proposes a special opacity caching strategy to accelerate the volume rendering.
Instant-NVR models the appearance of each body part with separate NeRFs and utilizes a carefully designed Chart-based backward skinning to model the deformation.
We conduct comparisons on three datasets: ZJU-MoCap~\cite{peng2021neural}, People-Snapshot~\cite{peopelsnap}, and UBC-Fashion~\cite{zablotskaia2019dwnet}. Similar to InstantAvatar~\cite{jiang2023instantavatar}, we also conduct test-time refinement of the SMPL pose.

\paragraph{ZJU-MoCap~\cite{peng2021neural}}
\begin{figure*}[h!]
\centering
   \includegraphics[width=1.0\textwidth]{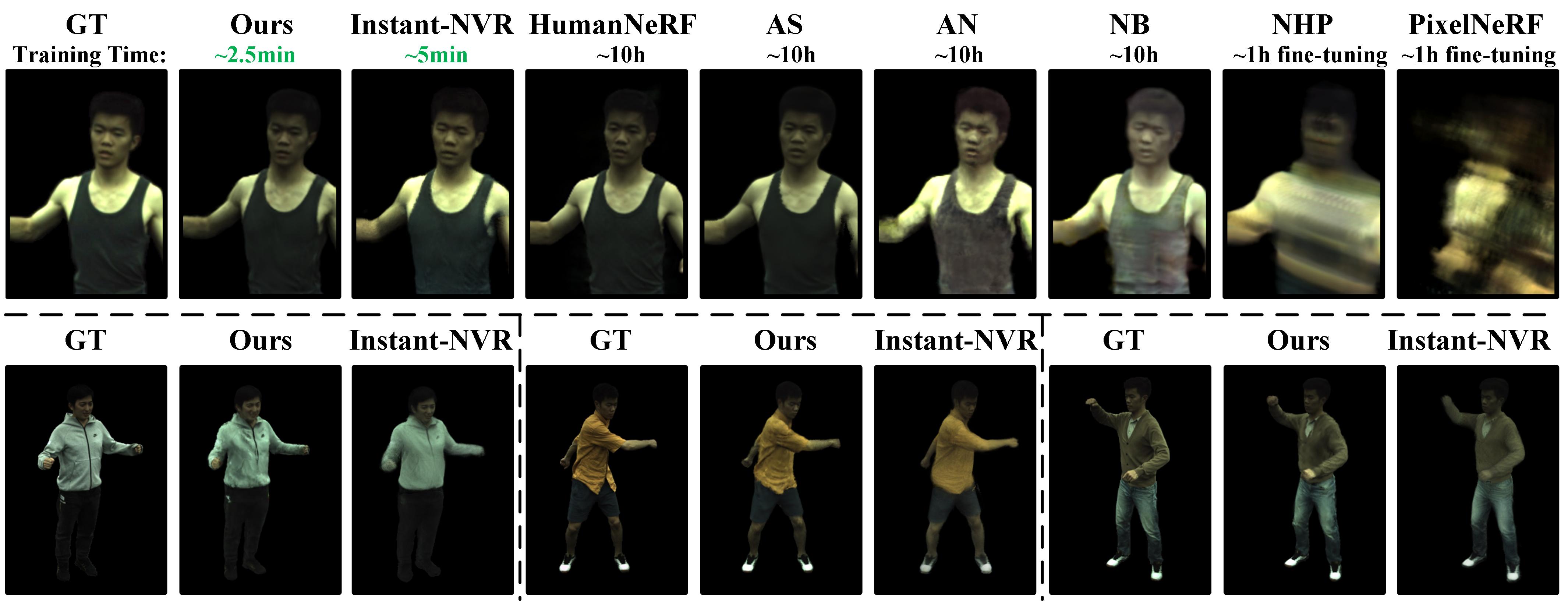}
    \caption{
    Comparison on ZJU-MoCap~\cite{peng2021neural}. The training time is highlighted under the method names.
    }
    \label{fig:zju}
\end{figure*}
\begin{table}[]
\centering
\scalebox{0.8}{
\begin{tabular}{@{}c|c|ccc@{}}
\toprule
Methods & Training time & PSNR & SSIM & LPIPS* \\ \midrule
HumanNeRF~\cite{weng2022humannerf} & $\sim$10h & 30.66 & 0.969 & 33.38 \\
AS~\cite{peng2022animatable} & $\sim$10h & 30.38 & 0.975 & 37.23 \\
NB~\cite{peng2021neural} & $\sim$10h & 29.03 & 0.964 & 42.47 \\
AN~\cite{peng2021animatable} & $\sim$10h & 29.77 & 0.965 & 46.89 \\
NHP~\cite{kwon2021neural} & $\sim$1h tuning & 28.25 & 0.955 & 64.77 \\
PixelNeRF~\cite{yu2021pixelnerf} & $\sim$1h tuning & 24.71 & 0.892 & 121.86 \\
Instant-NVR~\cite{instant_nvr} & $\sim$5min & 31.01 & 0.971 & 38.45 \\ \midrule
GART & $\sim$2.5min & \textbf{32.22} & \textbf{0.977} & \textbf{29.21} \\
GART & $\sim$30s & 31.76 & 0.976 & 34.01 \\ \bottomrule
\end{tabular}
}
\vspace{-2mm}
\caption{Comparison of view synthesis on ZJU-MoCap~\cite{peng2021neural}.}
\label{tab:zju}
\end{table}
We compare with Instant-NVR~\cite{instant_nvr} and other human rendering methods on the ZJU-MoCap dataset~\cite{peng2021neural} with the same setup as \cite{instant_nvr}. The average results are shown in Tab.~\ref{tab:zju}. {\Ours} surpasses other methods in terms of synthesis results with less training time. Thanks to its efficient rendering~\cite{kerbl20233d} and forward skinning (Sec.~\ref{sec:motion}), {\Ours} can achieve similar quantitative performance after less than $30$ seconds of training. Qualitative results in Fig.~\ref{fig:zju} show that our results capture more details than Instant-NVR~\cite{instant_nvr}.

\paragraph{People-Snapshot~\cite{peopelsnap}} 
\begin{figure*}[h!]
\centering
   \includegraphics[width=1.0\textwidth]{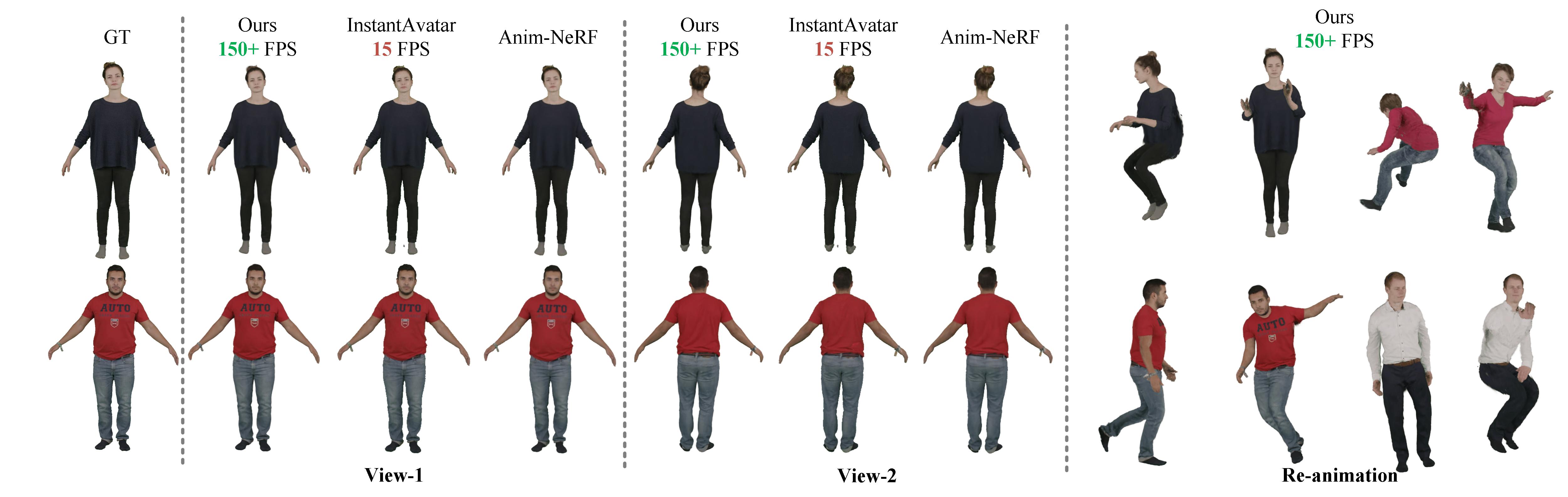}
    \caption{Comparison on People-Snapshot~\cite{peopelsnap}. Note our method achieves similar quality via shorter training time and $10\times$ inference FPS.}
    \label{fig:people}
\end{figure*}
\begin{table*}[]
\centering
\scalebox{0.8}{
\begin{tabular}{@{}
>{\columncolor[HTML]{FFFFFF}}c |
>{\columncolor[HTML]{FFFFFF}}c 
>{\columncolor[HTML]{FFFFFF}}c 
>{\columncolor[HTML]{FFFFFF}}c |
>{\columncolor[HTML]{FFFFFF}}c 
>{\columncolor[HTML]{FFFFFF}}c 
>{\columncolor[HTML]{FFFFFF}}c |
>{\columncolor[HTML]{FFFFFF}}c 
>{\columncolor[HTML]{FFFFFF}}c 
>{\columncolor[HTML]{FFFFFF}}c |
>{\columncolor[HTML]{FFFFFF}}c 
>{\columncolor[HTML]{FFFFFF}}c 
>{\columncolor[HTML]{FFFFFF}}c @{}}
\toprule
 & \multicolumn{3}{c|}{\cellcolor[HTML]{FFFFFF}male-3-casual} & \multicolumn{3}{c|}{\cellcolor[HTML]{FFFFFF}male-4-casual} & \multicolumn{3}{c|}{\cellcolor[HTML]{FFFFFF}female-3-casual} & \multicolumn{3}{c}{\cellcolor[HTML]{FFFFFF}female-4-casual} \\ \midrule
 & PSNR & SSIM & LPIPS & PSNR & SSIM & LPIPS & PSNR & SSIM & LPIPS & PSNR & SSIM & LPIPS \\
Neural Body~\cite{peng2021neural} ($\sim$14h) & 24.94 & 0.9428 & 0.0326 & 24.71 & 0.9469 & 0.0423 & 23.87 & 0.9504 & 0.0346 & 24.37 & 0.9451 & 0.0382 \\
Anim-NeRF~\cite{chen2021animatable} ($\sim$13h) & 29.37 & 0.9703 & \textbf{0.0168} & 28.37 & 0.9605 & \textbf{0.0268} & \textbf{28.91} & \textbf{0.9743} & \textbf{0.0215} & 28.90 & 0.9678 & \textbf{0.0174} \\
Anim-NeRF~\cite{chen2021animatable} (1m) & 12.39 & 0.7929 & 0.3393 & 13.10 & 0.7705 & 0.3460 & 11.71 & 0.7797 & 0.3321 & 12.31 & 0.8089 & 0.3344 \\
InstantAvatar~\cite{jiang2023instantavatar} (1m) & 29.65 & 0.9730 & 0.0192 & \textbf{27.97} & 0.9649 & 0.0346 & 27.90 & 0.9722 & 0.0249 & 28.92 & 0.9692 & 0.0180 \\ \midrule
GART (30s) & \textbf{30.40} & \textbf{0.9769} & 0.0377 & 27.57 & \textbf{0.9657} & 0.0607 & 26.26 & 0.9656 & 0.0498 & \textbf{29.23} & \textbf{0.9720} & 0.0378 \\ \bottomrule
\end{tabular}
}
\vspace{-2mm}
\caption{Comparison on People-Snapshot~\cite{peopelsnap}. InstantAvatar~\cite{jiang2023instantavatar} can inference at $15$ FPS while {\Ours} achieves $150+$ FPS.}
\label{tab:people}
\end{table*}
Another commonly compared human avatar dataset is People-Snapshot~\cite{peopelsnap}, and we compare {\Ours} to Instant-Avatar~\cite{jiang2023instantavatar} with the same experimental setup. The results are shown in Tab.~\ref{tab:people} and Fig.~\ref{fig:people}. Our method achieves comparable performance with shorter training time. Besides the training efficiency, {\Ours} has a unique advantage over Instant-Avatar during inference. At the resolution of $540\times 540$, Instant-Avatar can be rendered at $15$FPS~\cite{jiang2023instantavatar}, but {\Ours} can be rendered at more than $150$FPS on a single RTX-3080-Laptop GPU.

\paragraph{UBC-Fashion~\cite{zablotskaia2019dwnet}}
\begin{figure*}[t!]
\centering
   \includegraphics[width=1.0\textwidth]{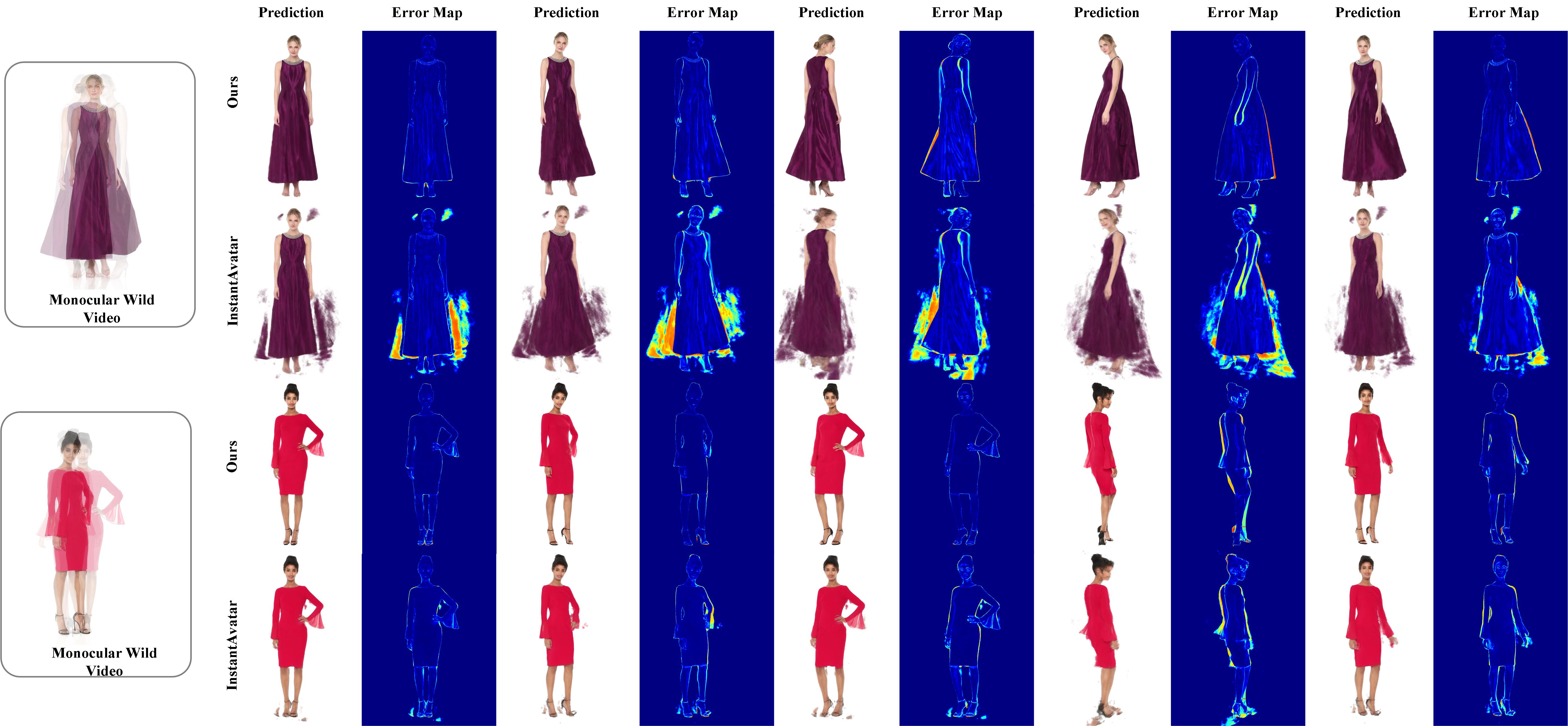}
    \caption{Comparison on UBC-Fashion~\cite{zablotskaia2019dwnet} challenging sequences. Note how the SoTA method~\cite{jiang2023instantavatar} makes artifacts around the long skirts (top) and feet (bottom).} 
    \label{fig:ubc}
    \vspace{-1.5em}
\end{figure*}
\begin{table}[]
\begin{tabular}{@{}c|ccc@{}}
\toprule
Methods & PSNR & SSIM & LPIPS* \\ \midrule
InstantAvatar~\cite{jiang2023instantavatar} Test Pose & 19.87 & 0.880 & 157.70 \\
InstantAvatar~\cite{jiang2023instantavatar} Training Pose & 19.97 & 0.882 & 157.05 \\
GART MLP & 25.65 & 0.934 & 81.88 \\
GART T-Table & \textbf{25.96} & \textbf{0.935} & \textbf{80.57} \\ \bottomrule
\end{tabular}
\vspace{-2mm}
\caption{Quantitative comparison of view synthesis on UBC-Fashion~\cite{zablotskaia2019dwnet} sequences.}
\label{tab:ubc}
\end{table}
While the ZJU-MoCap~\cite{peng2021neural} and People-Snapshot~\cite{peopelsnap} datasets are widely benchmarked, the clothing in these datasets is all tight and does not differ much from the SMPL body model. We take a step forward towards modeling more challenging clothing, such as long dresses with highly dynamic motion and deformation. We use six videos from the UBC-Fashion~\cite{zablotskaia2019dwnet} dataset that contains dynamic dresses and different skin colors. As shown in Fig.~\ref{fig:ubc}, each monocular video captures a model wearing loose clothing in front of the camera and turning around. 
Since these sequences have very limited variation of poses and only capture one view, we use the frames starting from $0$ and pick frames with an interval of $4$ for training and use the frames starting from $2$ and pick frames with the same interval for testing. The SMPL poses are obtained via the SoTA human pose estimator ReFit~\cite{wang2023refit}. 
Because both {\Ours} and Instant-Avatar~\cite{jiang2023instantavatar} can optimize the SMPL pose during testing, and the pose estimation is noisy since the long skirts also lead to challenges for pose estimators, we found that the results of both methods are better if we use the nearest training pose and optimize it during testing.

The quantitative comparison is shown in Tab.~\ref{tab:ubc}, where we evaluate two variants of {\Ours}: {\Ours}-MLP uses an MLP to represent the latent bones $\widetilde{\mathcal{B}}(\theta)$ in Eq.~\ref{eq:add_bones}, where the input to the MLP is the SMPL Pose; {\Ours}-T-Table directly optimize a list of rigid transformations per-time-frame that represents the latent bones $[\widetilde B_{1},\ldots, \widetilde B_{n_l}]$. 
As shown in Fig~\ref{fig:ubc}, Instant-Avatar successfully captures the upper body but fails to capture the dynamic clothing.
There are three potential reasons: 1) Because of the implicit modeling, Fast-SNARF~\cite{fastsnarf} is utilized to solve the backward skinning, leading to multiple ambiguous correspondences in the highly dynamic skirt area. So we observe the wrong skinning that attaches skirts to the arm. 2) Using 24 SMPL bones and learnable skinning weights is insufficient to capture the complex deformation; 3) Because of the limited expressiveness of the deformation but the flexible nature of NeRF and the noisy pose estimation, many artifacts are created in the empty space due to their photometric significance at some wrong poses.
On the contrary, our deformation is modeled via simple forward skinning, which can further capture flexible deformation via latent bones as in Sec.~\ref{sec:motion}, and is optimized with 3D-GS~\cite{kerbl20233d} in a deformation-based process, which leads to our better performance.

\vspace{\vspacebeforesection}
\subsection{Application on Dog Rendering}
\vspace{\vspaceaftersection}
\begin{figure*}[t!]
\centering
   \includegraphics[width=1.0\textwidth]{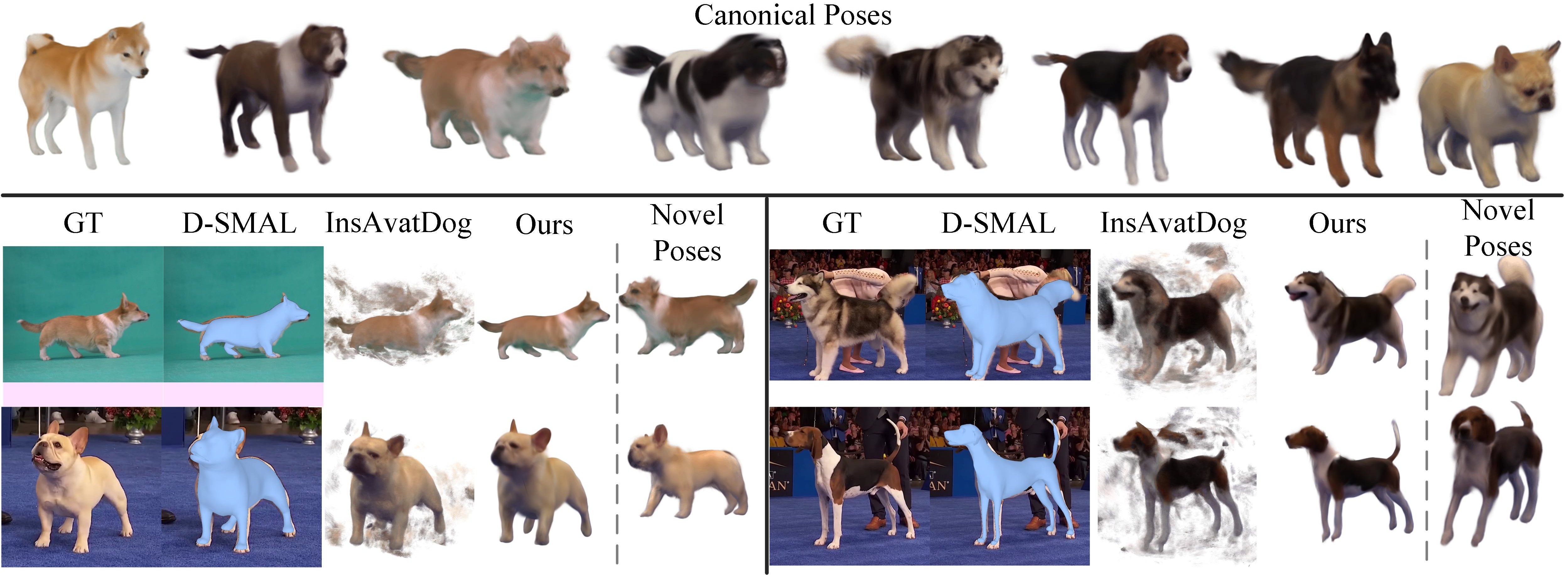}
    \caption{Qualitative results for in-the-wild dogs in canonical pose (Top) and in novel poses and views (Bottom).}
    \label{fig:dog}
    \vspace{-1em}
\end{figure*}
\begin{table}[]
\centering
\begin{tabular}{@{}c|c|ccc@{}}
\toprule
Data & Method & PSNR & SSIM & LPIPS \\ \midrule
\multirow{2}{*}{\shortstack{National Dog \\Show (6 seq)} }
    & InsAvat-Dog & 16.13 & 0.759 & 0.318 \\
    & {\Ours} & \textbf{17.86} & \textbf{0.825} & \textbf{0.238} \\
\midrule
\multirow{2}{*}{\shortstack{Adobe Stock \\(2seq)}}
    & InsAvat-Dog & 20.62 & 0.834 & 0.227 \\
    & {\Ours} & \textbf{24.50} & \textbf{0.921} & \textbf{0.114} \\
\bottomrule
\end{tabular}
\vspace{-2mm}
\caption{Quantitative evaluation of view synthesis on ITW dogs.}
\label{tab:dog}
\end{table}
In this section, we demonstrate {\Ours} as a general framework to capture and render animals from monocular in-the-wild videos. Specifically, we utilize the new D-SMAL~\cite{ruegg2023bite} model that is proposed for diverse dog breeds as the base template. We conduct experiments on a total of 8 new sequences: 6 sequences from the 2022 National Dog Show (the 6 best-in-show participants), and 2 sequences captured with a green screen obtained from Adobe Stock Videos. 
Compared to humans, pose estimation for dogs is more challenging because of the scarcity of training data and occlusions in the environment. Therefore, we select sections where the poses are estimated corrected by BITE~\cite{ruegg2023bite}, and there are few occlusions. 
As shown in Fig.~\ref{fig:dog}, {\Ours} captures different dog breeds well. 
Compared to D-SMAL, {\Ours} better reconstructs breed-specific appearance such as tails, ears, and textured fur. 
We also adapt InstantAvatar~\cite{jiang2023instantavatar} to the D-SMAL template, which we call InsAvat-Dog for comparison. Compared to {\Ours}, InsAvat-Dog may create ghost artifacts under such a challenging setting, potentially due to the inaccurate and highly dynamic dog poses during training.
We include a small set of test frames for each sequence and report the metrics in Tab.~\ref{tab:dog} as a baseline for neural animal reconstruction in the wild.

\vspace{\vspacebeforesection}
\subsection{Ablation Study}
\vspace{\vspaceaftersection}
\begin{figure*}[t!]
\centering
   \includegraphics[width=1.0\textwidth]{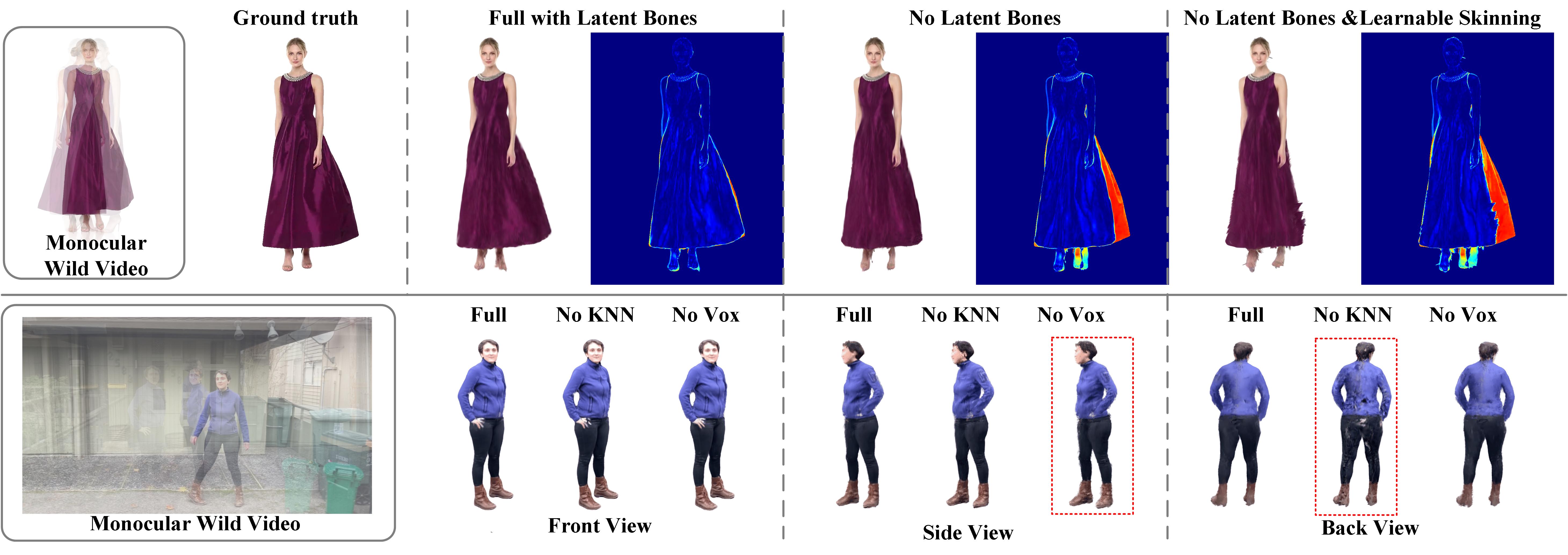}
    \caption{\textbf{Ablation}: (Top) Learnable skinning and latent bones help {\Ours} to capture highly dynamic skirts. (Bottom) KNN regularization in Eq.~\ref{eq:knn_reg} helps smooth the results in the back view, and voxel-distilled skinning helps prevent noisy artifacts on the side view.}
    \label{fig:abl}
    \vspace{-1em}
\end{figure*}
\begin{table}[]
\centering
\begin{tabular}{@{}c|ccc@{}}
\toprule
Methods & PSNR & SSIM & LPIPS* \\ \midrule
No Learnable Skinning & 23.76 & 0.925 & 88.76 \\
No Latent Bones & 25.00 & 0.932 & 82.03 \\
Full & 25.65 & 0.934 & 81.88 \\ \bottomrule
\end{tabular}
\vspace{-2mm}
\caption{Ablation of learned deformation on UBC-Fashion~\cite{zablotskaia2019dwnet}.}
\label{tab:abl_deform}
\end{table}
To verify the effectiveness of our deformation modeling, we compare the full model with 1) removing the latent bones and 2) removing the learnable skinning on the UBC-Fashion sequences. The results are shown in Tab.~\ref{tab:abl_deform} and Fig.~\ref{fig:abl}. 
We observe that the full model works the best and note that the model without latent bones can still reconstruct the dress with fewer artifacts than Instant-Avatar~\cite{jiang2023instantavatar}, showing our robustness and effectiveness under noisy poses and large deformations. Visually, we observe from Fig.~\ref{fig:abl} that when adding the latent bones, the independent motion of the skirts can be captured better than the ablated ones.
We also verify the effectiveness of our injected smoothness by 1) removing the voxel distilled skinning weight but storing the skinning weight for each Gaussian as a list, and 2) removing the KNN regularization as in Eq.~\ref{eq:knn_reg}. The qualitative comparisons of in-the-wild video are shown in Fig.~\ref{fig:abl}. 
We note that the No-KNN version results in strong artifacts on the back, while the No-Vox version produces noisy artifacts around the body in the side view.

\subsection{Application: Text-to-GART}
\begin{figure*}[t!]
\centering
   \includegraphics[width=1.0\textwidth]{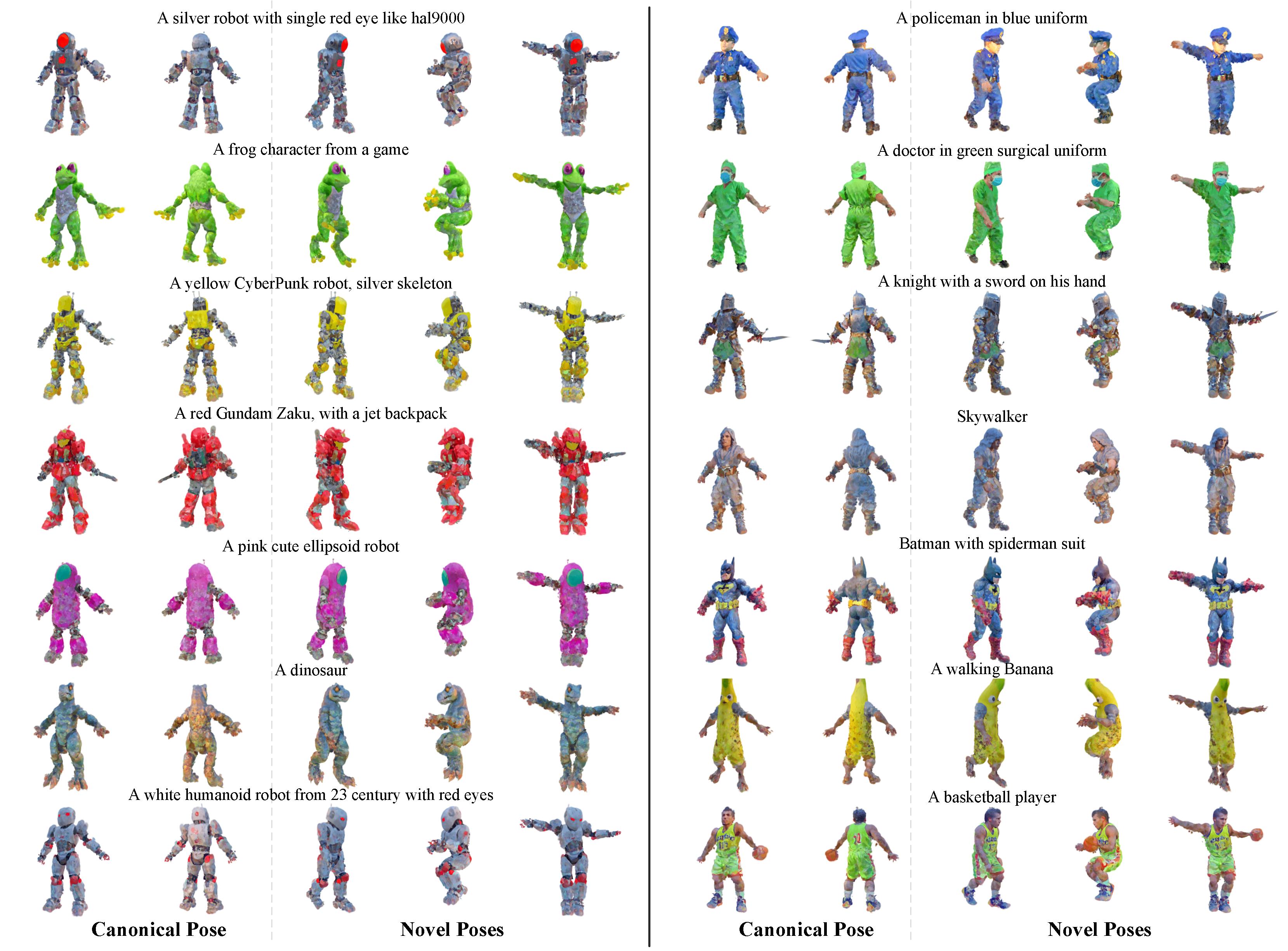}
    \caption{\textbf{Additional application}: Text-to-GART
    }
    \label{fig:text2gart}
    \vspace{-1em}
\end{figure*}
{\Ours} is a general representation of articulated subjects and is not restricted to real monocular video reconstruction. In this section, we further demonstrate an application -- Text-to-GART, by simply changing the rendering $L1$ loss and SSIM loss in Eq.~\ref{eq:total_loss} to an SDS loss~\cite{poole2022dreamfusion}. The input is a text describing the content the user aims to generate, and the output is an optimized {\Ours} representing this subject.
The optimization loss becomes
$L = L_{\text{SDS}}+ L_{\text{reg}}$,
where $L_{\text{SDS}}$ is computed via forwarding a fine-tuned Stable-Diffusion~\cite{rombach2022high} model MVDream~\cite{shi2023MVDream}. For more details on $L_{\text{SDS}}$, please see Stable-Diffusion~\cite{rombach2022high} and DreamGaussian~\cite{tang2023dreamgaussian}.
Since there are no real poses estimated from video frames, we randomly sample some reasonable SMPL~\cite{smpl} template poses from AMASS~\cite{AMASS:ICCV:2019} to augment {\Ours} during distillation.
The generation results are shown in Fig.~\ref{fig:text2gart}.
Thanks to the efficiency of {\Ours}, the computation bottleneck of this application is mainly in the 2D diffusion forwarding, and the typical generation time is around 10 minutes per subject on a single GPU.
\vspace{\vspacebeforesection}
\section{Conclusions}
\vspace{\vspaceaftersection}
This paper proposes a simple and general representation, {\Ours}, for non-rigid articulated subjects via Gaussian Mixtures and novel skinning deformation. {\Ours} achieves SoTA performance on monocular human and animal reconstruction and rendering while maintaining high training and inference efficiency.
\paragraph{Limitations and future work}
Our proposed method has two main limitations, which could be explored in the future: 
1) Our method relies on a template pose estimator, which may not exist for more general animal species.
2) {\Ours} can fit a single monocular video efficiently, and it's an interesting next step to explore how to capture the category-level prior knowledge of articulated subjects from in-the-wild video collections.

\paragraph{Acknowledgements} The authors appreciate the support of the following grants: NSF NCS-FO 2124355, NSF FRR 2220868.

\newpage
{\small
\bibliographystyle{ieee_fullname}
\bibliography{reference.bib}
}

\end{document}